%% file: ijcai25.tex
\documentclass{article}
\usepackage{ijcai25}

\usepackage{times}
\usepackage{soul}
\usepackage{url}
\usepackage[hidelinks]{hyperref}
\usepackage[utf8]{inputenc}
\usepackage[small]{caption}
\usepackage{graphicx}
\usepackage{amsmath}
\usepackage{amsthm}
\usepackage{amssymb}
\usepackage{booktabs}
\usepackage{algorithm}
\usepackage{algorithmic}
\usepackage[switch]{lineno}
\usepackage{bm}
\usepackage{bbm}
\usepackage{xspace}
\usepackage{pifont}
\usepackage{color}
\usepackage{colortbl}
\usepackage{xcolor}
\usepackage{array}
\usepackage{multirow}
\usepackage{soul} 

\definecolor{myPink}{RGB}{247, 219, 231}  
\definecolor{myGreen}{RGB}{227, 241, 217}   
\definecolor{myBlue}{RGB}{187, 230, 240}    
\definecolor{myYellow}{RGB}{255, 242, 204}    
\definecolor{myOrange}{RGB}{255, 229, 204}    
\definecolor{myPurple}{RGB}{224, 208, 250}    

\newcommand{\hlpink}[1]{\sethlcolor{myPink}\hl{#1}}
\newcommand{\hlgreen}[1]{\sethlcolor{myGreen}\hl{#1}}
\newcommand{\hlblue}[1]{\sethlcolor{myBlue}\hl{#1}}
\newcommand{\hlyellow}[1]{\sethlcolor{myYellow}\hl{#1}}

\newcommand{\hlpurple}[1]{\sethlcolor{myPurple}\hl{#1}}

\newcommand{\cmark}{\ding{51}}%
\def \ie {\emph{i.e.}}
\def \eg {\emph{e.g.}}

\definecolor{Gray}{gray}{0.9}
\definecolor{LightCyan}{rgb}{0.88,0.95,1}
\definecolor{TitleColor}{gray}{0.95}
\newcommand{\tit}[1]{\smallbreak\noindent\textbf{#1.}}

\title{Image Captioning Evaluation in the Age of Multimodal LLMs:\\Challenges and Future Perspectives}

\author{
Sara Sarto$^1$\and
Marcella Cornia$^1$\and
Rita Cucchiara$^{1,2}$\\
\affiliations
$^1$University of Modena and Reggio Emilia, Italy\\
$^2$IIT-CNR, Italy\\
\emails
\{sara.sarto, marcella.cornia, rita.cucchiara\}@unimore.it
\vspace{0.1cm}
\\
\footnotesize{\texttt{\url{https://github.com/aimagelab/awesome-captioning-evaluation}}}
}

\begin{document}

\maketitle
\begin{abstract}
The evaluation of machine-generated image captions is a complex and evolving challenge. With the advent of Multimodal Large Language Models (MLLMs), image captioning has become a core task, increasing the need for robust and reliable evaluation metrics. This survey provides a comprehensive overview of advancements in image captioning evaluation, analyzing the evolution, strengths, and limitations of existing metrics. We assess these metrics across multiple dimensions, including correlation with human judgment, ranking accuracy, and sensitivity to hallucinations. Additionally, we explore the challenges posed by the longer and more detailed captions generated by MLLMs and examine the adaptability of current metrics to these stylistic variations. Our analysis highlights some limitations of standard evaluation approaches and suggests promising directions for future research in image captioning assessment.
\end{abstract}

\sloppy

\input{sections/01_introduction}
\input{sections/02_captioning}

\input{sections/03_experiments}
\input{sections/04_future_directions}
\input{sections/05_conclusions}

\clearpage
\appendix


\section*{Acknowledgments}
We acknowledge the CINECA award under the ISCRA initiative, for the availability of high-performance computing resources. This work has been conducted under a research grant co-funded by Leonardo S.p.A. and supported by the PNRR-M4C2 project FAIR and the PRIN 2022-PNRR project MUCES (CUP E53D23016290001), funded by EU - Next-Generation EU. The authors also thank Davide Caffagni and Lorenzo Baraldi for their valuable support.

\bibliographystyle{named}
\bibliography{ijcai25}

\end{document}

%% file: sections/01_introduction.tex
\section{Introduction}\label{sec:intro}

Image captioning, the task of generating natural language descriptions of visual content~\cite{stefanini2022show}, is a fundamental component of modern Artificial Intelligence systems such as Multimodal Large Language Models (MLLMs) and autonomous agents. These models, indeed, need to understand and interact with visual data effectively and be capable of generating high-quality descriptions. As MLLMs continue to advance, the development of robust and reliable evaluation metrics for image descriptions becomes essential.

The history of image description has been marked by continuous innovation. Early captioning models evolved from template-based and recurrent neural networks~\cite{vinyals2015show} to Transformer-based architectures~\cite{cornia2020meshed,pan2020x}. Typically, an image captioning model consists of a visual encoder and a language model. Visual encoders range from CNNs~\cite{anderson2018bottom} to Vision Transformers~\cite{dosovitskiy2020image}, while language models have transitioned from LSTMs to Transformers~\cite{vaswani2017attention}. More recently, MLLMs, such as LLaVA~\cite{liu2024improved}, IDEFICS~\cite{laurenccon2023obelisc}, and Llama 3.2 Vision~\cite{dubey2024llama}, have integrated these capabilities, transforming how captions are generated and evaluated.

Training strategies have also evolved, incorporating supervised learning integrating both cross-entropy loss and reinforcement learning using self-critical sequence training~\cite{rennie2017self}. Several benchmark datasets have facilitated the development of image captioning models, including COCO~\cite{lin2014microsoft}, Conceptual Captions~\cite{sharma2018conceptual}, and nocaps~\cite{agrawal2019nocaps}, each providing varying levels of diversity and complexity.

As captioning models improve, the role of evaluation metrics becomes increasingly critical. A good metric must assess captions across multiple dimensions to ensure alignment with human expectations both linguistically and semantically. A strong captioning metric should reward captions that are \textbf{fluent}, \textbf{accurate}, and \textbf{faithful} to the image content, while also ensuring they focus on the \textbf{most salient aspects} and strike a balance between \textbf{detail} and \textbf{conciseness}. Furthermore, it should penalize hallucinated or misleading information.

\begin{figure}[t]
\includegraphics[width=0.99\linewidth]{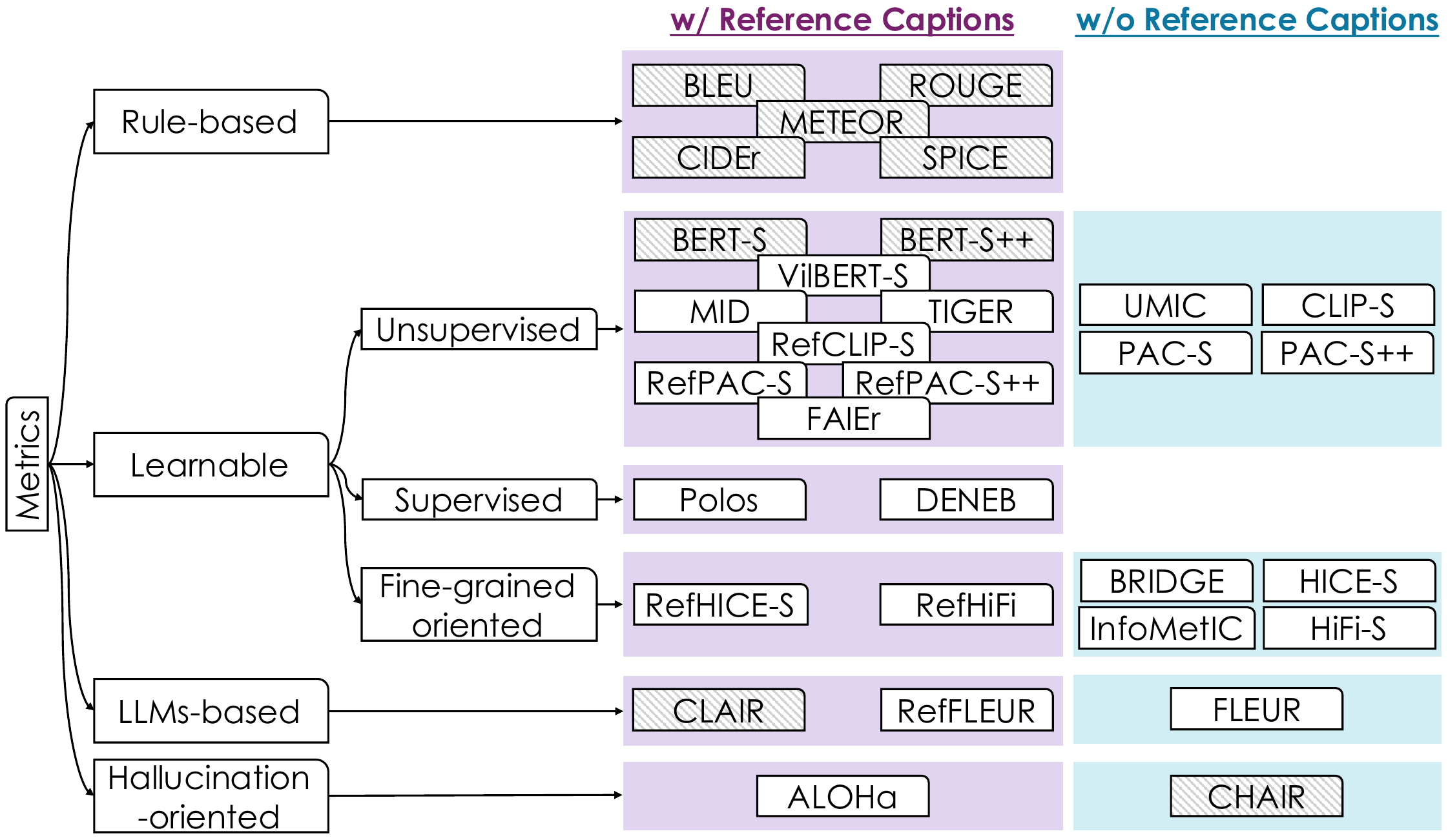}
\vspace{-0.1cm}
\caption{Taxonomy of image captioning metrics, distinguishing between reference-based and reference-free approaches. Grey dashed lines indicate the absence of the input image during evaluation.}
\label{fig:taxonomy}
\vspace{-0.3cm}
\end{figure}

Early metrics such as BLEU~\cite{papineni2002bleu}, ROUGE~\cite{lin2004rouge}, and METEOR~\cite{banerjee2005meteor} were borrowed from machine translation and focused on $n$-gram overlap with reference captions. Later, specialized metrics like CIDEr~\cite{vedantam2015cider} and SPICE~\cite{spice2016} aimed to improve correlation with human judgment by incorporating consensus-based scoring and scene graph analysis, respectively. While these metrics introduced refinements, they remained constrained by their reliance on reference captions and linguistic overlap.

Recent advances have shifted towards learnable metrics leveraging vision-and-language pre-trained models. CLIP-S~\cite{hessel2021clipscore} introduces a contrastive approach to measure image-caption alignment based on CLIP embeddings. Fine-grained metrics like BRIDGE~\cite{sarto2024bridge} and HiFi-S~\cite{yao2024hifi} capture localized semantics and hierarchical structures. LLM-based metrics such as CLAIR~\cite{chan2023clair} and FLEUR~\cite{lee2024fleur} integrate reasoning and explanation into caption evaluation.

With the rise of MLLMs as image description generators, the nature of captions has changed, as these models generate captions that differ in length, specificity, and abstraction. Conventional evaluation metrics may no longer reliably assess caption quality in this new paradigm, posing new challenges to the evaluation of generated captions. Following this insight, this survey provides a comprehensive analysis of existing captioning evaluation metrics, from traditional rule-based approaches to modern learnable methods and LLM-driven techniques. We explore their evolution, strengths, and limitations while emphasizing the growing need for robust evaluation frameworks that can adapt to the stylistic diversity introduced by MLLMs. Additionally, we present an in-depth experimental study assessing metrics across multiple dimensions, including correlation with human judgment, ranking accuracy, and sensitivity to hallucinations. Our analysis highlights the performance gaps of conventional metrics and showcases promising directions for future research.

%% file: sections/02_captioning.tex
\section{The Evolution of Captioning Metrics}\label{sec:captioning}
In this section, we present the evolution of captioning metrics, outlining their development from traditional non-learning-based evaluation strategies to recent advancements incorporating LLMs. Fig.~\ref{fig:taxonomy} presents a taxonomy that highlights the key characteristics of the most commonly used metrics, including their reliance on reference captions or image input.

\subsection{Rule-based Metrics}
Many widely used evaluation metrics for image captioning originate from NLP tasks and rely heavily on $n$-gram matching techniques. Popular metrics in this domain include BLEU~\cite{papineni2002bleu}, METEOR~\cite{banerjee2005meteor}, ROUGE~\cite{lin2004rouge}, CIDEr~\cite{vedantam2015cider}, and SPICE~\cite{spice2016}. Despite their extensive use in image captioning, these traditional metrics exhibit several limitations. Notably, they do not consider the input image during evaluation and often demonstrate weak correlations with human judgment. Also, their reliance on reference captions restricts their ability to comprehensively evaluate generated captions, as their effectiveness is inherently influenced by the style and quality of the reference dataset.

BLEU~\cite{papineni2002bleu} and METEOR~\cite{banerjee2005meteor} were originally designed for machine translation tasks. BLEU evaluates $n$-gram precision and introduces the concept of modified unigram precision, which adjusts word frequency counts by considering the maximum occurrence of each word across all reference translations.
By design, it penalizes overly long translations through precision-based scoring and mitigates excessively short ones with a brevity penalty, promoting balance in word choice, order, and length.
However, BLEU lacks a direct measure of recall, which quantifies the proportion of matched $n$-grams in the reference. Recall is particularly crucial for evaluating translation quality, as it reflects the extent to which a candidate text captures the content of the reference. Introduced to mitigate this limitation, METEOR prioritizes recall by considering unigram matches between candidate and reference sentences and incorporating exact matches, stemmed forms, and semantic equivalences. This enables METEOR to provide a more detailed and holistic assessment of translation quality. Similarly, ROUGE~\cite{lin2004rouge}, initially developed for summarization tasks, has been adapted for evaluating visual descriptions and emphasizes $n$-gram recall by analyzing the overlap of $n$-grams between candidate and reference texts. 

Recognizing the limitations of these general-purpose metrics, CIDEr~\cite{vedantam2015cider} and SPICE~\cite{spice2016} were designed for the captioning task. CIDEr employs a TF-IDF weighting scheme to reweight the significance of different $n$-grams, capturing both precision and recall while accounting for their contextual importance. SPICE takes a structured approach by converting captions into scene graphs and measuring their similarity, aligning the evaluation with the semantic and structural elements of the captions.

\subsection{Learnable Metrics}
All the aforementioned metrics primarily rely on textual-level comparisons and struggle to adequately account for synonym matches from a linguistic perspective. Moreover, they assume that human-written reference captions perfectly reflect the content of the image, which may not always be the case. To overcome these limitations, recent metrics leverage pre-trained models to compare textual-only and visual-textual content, using both supervised and unsupervised approaches.

\tit{Unsupervised Metrics}
BERT-S~\cite{zhang2019bertscore}, for instance, utilizes learned embeddings from the pre-trained language model BERT~\cite{devlin2018bert} to more effectively measure semantic similarities between candidate and reference captions. Building on this, BERT-S++~\cite{yi2020improving} enhances the approach by incorporating the variance across multiple reference captions, improving robustness.
However, these metrics still do not integrate the visual modality, which can limit their ability to fully evaluate the performance of captioning models. To overcome this, metrics that explicitly include the image as input have been introduced. These methods leverage pre-trained vision-and-language models, such as UNITER~\cite{chen2020uniter}, ViLBERT~\cite{lu2019vilbert}, and CLIP~\cite{radford2021learning}, to evaluate captions in a way that aligns more closely with both textual and visual content.

Among these metrics, UMIC~\cite{lee2021umic} stands out as it effectively addresses a key limitation of BERT-S: the difficulty of evaluating image captions in the absence of sufficient reference captions. To overcome this, it introduces a reference-free approach built on a fine-tuned version of a pre-trained multimodal Transformer (\ie, UNITER). By leveraging contrastive learning, the model is trained to differentiate between ground-truth captions and diverse synthetic negative samples. These negative samples are meticulously designed to cover various undesirable captioning cases, such as those that are grammatically incorrect, irrelevant to the image, or relevant but containing incorrect keywords. This approach ensures a more robust evaluation of image captioning models without relying on reference captions.

TIGEr~\cite{jiang2019tiger}, on the other hand, focuses on the integration of the image as one of the inputs for computing an evaluation score. Building upon a text-image grounding model~\cite{lee2018stacked}, it allows evaluating caption quality not only based on how well a caption represents image content but also on how well machine-generated captions match human-generated ones.
Following this line and inspired by BERT-S, ViLBERT-S~\cite{lee2020vilbertscore} computes cosine similarity between token embeddings for reference and candidate sentences. However, different from BERT-S,
the token embedding is computed with the consideration of image contexts.
On a different line, FAIEr~\cite{wang2021faier} employs the scene graph as a bridge between the image and textual modality. Specifically, it fuses scene
graphs of the image and references as a union scene graph and compares it with the scene graph of generated captions.

Leveraging the success of employing pre-trained vision-and-language models, numerous metrics started exploiting the CLIP extensive pre-training to compute image captioning evaluation scores. Among these, CLIP-S~\cite{hessel2021clipscore} was the first metric to utilize a modified cosine similarity between image and candidate caption representations derived from the visual and textual encoders of CLIP. In a similar way, MID~\cite{kim2022mutual} leverages CLIP-based visual-textual features to compute negative Gaussian cross-mutual information, yielding a more effective evaluation metric.

While these advancements highlight the potential of contrastive-based embedding spaces for caption evaluation, large-scale models pre-trained on web-sourced data have inherent limitations. In particular, CLIP, which is trained on web-collected image-caption pairs, may be suboptimal for image evaluation tasks, as such annotations often lack the richness, grammatical correctness, and detail required to assess generated, typically longer captions. To address this, methods like PAC-S~\cite{sarto2023positive} and its improved version PAC-S++~\cite{sarto2024positive} aim to refine the CLIP embedding space by incorporating a fine-tuning phase. This process incorporates curated image-caption pairs along with synthetically generated positive examples, achieving improved performance in image captioning evaluation.

\tit{Supervised Metrics}
All the aforementioned metrics leverage the CLIP embedding space, either pre-trained or fine-tuned on image-caption pairs. However, CLIP was not originally designed specifically for computing evaluation scores, which can limit its effectiveness. To address this challenge, a less explored research direction focuses on refining the CLIP embedding space through supervised training approaches. For instance, the Polos metric~\cite{wada2024polos} is based on Polaris, a dataset specifically designed for predicting evaluation scores under direct supervision from human-annotated judgments. As part of this work, the authors propose a multimodal metric learning framework based on human feedback, which handles both image and text inputs and learns directly from human evaluations based on multimodal inputs.

Building on this foundation, the DENEB metric~\cite{matsuda2024deneb} introduces a supervised evaluation approach specifically designed to be robust against hallucinations. DENEB processes multiple reference captions simultaneously to effectively capture similarities between an image, a candidate caption, and reference captions. Furthermore, to improve the visual diversity of the Polaris dataset, DENEB introduces Nebula, an extended dataset containing approximately three times the number of images, further enhancing its robustness and scalability for evaluation tasks.

\begin{table*}[t]
\small
\centering
\setlength{\tabcolsep}{.2em}
\resizebox{\linewidth}{!}{
\begin{tabular}{lc ccc cc ccc cc cc cc cc c }
\toprule
 &  &  & \multicolumn{2}{c}{\textbf{Inputs}} & & \multicolumn{1}{c}{\cellcolor{myPink}{\textbf{Flickr8k-Expert}}} &  \cellcolor{myPink}{} & \multicolumn{1}{c}{\cellcolor{myPink}{\textbf{Flickr8k-CF}}} & \cellcolor{myPink}{}  & \multicolumn{1}{c}{\cellcolor{myPink}{\textbf{Polaris}}} & \cellcolor{myPink}{} & \multicolumn{1}{c}{\cellcolor{myPink}{\textbf{Nebula}}} & \cellcolor{myPink}{} & \multicolumn{1}{c}{\cellcolor{myPink}{\textbf{Composite}}} &  & \multicolumn{1}{c}{\cellcolor{myBlue}{\textbf{Pascal-50S}}} &  & \multicolumn{1}{c}{\cellcolor{myPurple}{\textbf{FOIL}}} \\
\cmidrule{4-5} \cmidrule{7-7} \cmidrule{9-9} \cmidrule{11-11} \cmidrule{13-13} \cmidrule{15-15} 
\cmidrule{17-17} \cmidrule{19-19}
& \textbf{External Models} & & \textbf{Image} & \textbf{Refs} && Kendall $\tau_c$ & & Kendall $\tau_b$ & & Kendall $\tau_c$ & & Kendall $\tau_c$ & & Kendall $\tau_c$ & & Accuracy & & Accuracy \\
\midrule
\rowcolor{TitleColor}\multicolumn{18}{l}{\textit{Rule-based}} & \\
BLEU-4~\cite{papineni2002bleu}  & - & &  & \cmark && 30.8 & & 16.9 && 46.3 && 40.4 && 30.6 && 74.0 && 66.2\\
ROUGE~\cite{lin2004rouge} & - & &  & \cmark && 32.3 & & 19.9 && 46.3 && 42.6 && 32.4 && 78.0 && 54.6\\
METEOR~\cite{banerjee2005meteor}  & - & &  & \cmark && 41.8 & & 22.2 && 51.2 && 46.8 && 38.9 && \textbf{81.1} && 70.1\\
CIDEr~\cite{vedantam2015cider}  & - & &  & \cmark && 43.9 & & \textbf{24.6} && \textbf{52.1} && \textbf{48.1} && 37.7 && 80.1 && \textbf{85.7} \\
SPICE~\cite{spice2016} & - & &  &  \cmark && \textbf{44.9} & & 24.4 && 51.0 && 44.0 && \textbf{40.3} && 76.7 && 75.5\\
\midrule
\rowcolor{TitleColor}\multicolumn{18}{l}{\textit{Learnable Unsupervised}} & \\
FAIEr~\cite{wang2021faier} & - & & \cmark & \cmark && - & & - && - && - && - && 81.4 && -\\
TIGEr~\cite{jiang2019tiger} & SCAN & & \cmark & \cmark && 49.3 & & - && - && - && 45.4 && 80.7 && -\\
UMIC~\cite{lee2021umic} & $\text{UNITER}_{\text{BASE}}$ & & \cmark &  && 46.8 & & - && 49.8 && - && 56.1 && 85.1 && -\\
BERT-S~\cite{zhang2019bertscore} & $\text{BERT}_{\text{BASE}}$ & &  & \cmark && 39.2 & & 22.8 && 51.6 && 47.0 && 30.1 && 80.1 && 88.6\\
BERT-S++~\cite{yi2020improving} & $\text{BERT}_{\text{BASE}}$ & &  & \cmark && 46.7 & & - && - && - && 44.9 && 80.1 && -\\
ViLBERT-S~\cite{lee2020vilbertscore} & $\text{ViLBERT}_{\text{BASE}}$ & & \cmark & \cmark && 50.1 & & - && - && - && 52.4 && 79.6 && -\\
MID~\cite{kim2022mutual} & CLIP ViT-B & & \cmark & \cmark && 54.9 & & 37.3 && 51.3 && \textbf{51.3} && - && 85.2 && 90.5\\
CLIP-S~\cite{hessel2021clipscore} & CLIP ViT-B & & \cmark &  && 51.2 & & 34.4 && 52.3 && 46.9 && 53.8 && 80.9 && 87.2\\
RefCLIP-S~\cite{hessel2021clipscore} & CLIP ViT-B & & \cmark & \cmark && 53.0 & & 36.4 && 52.3 && 46.9 && 55.4 && 83.3 && 91.0 \\
PAC-S~\cite{sarto2023positive} & CLIP ViT-B & & \cmark &  && {54.3} & & {36.0} && 52.3 && 47.2 && {55.7} && 82.4 && 89.9\\
RefPAC-S~\cite{sarto2023positive} & CLIP ViT-B  & & \cmark & \cmark && {{55.9}} & & {{37.6}} && 55.2 && 50.6 && {{57.3}} && 84.7 && 93.7\\
PAC-S++~\cite{sarto2024positive} & CLIP ViT-B & & \cmark &  && {54.5} & & {37.0} && 52.4 && - && {58.3} && 82.3 && 90.2\\
RefPAC-S++~\cite{sarto2024positive} & CLIP ViT-B  & & \cmark & \cmark && {{55.7}} & & {{37.9}} && 54.8 && - && {{59.1}} && 84.5 && 93.5\\
InfoMetIC~\cite{hu2023infometic} & CLIP ViT-B & & \cmark &  && 55.5 & & 36.6 && - && - && 59.3 && \textbf{86.5} && -\\
CLIP-S~\cite{hessel2021clipscore} & CLIP ViT-L & & \cmark &  && 52.6 & & 35.2 && 53.2 && - && 55.4 && 81.7 && 90.9\\
RefCLIP-S~\cite{hessel2021clipscore} & CLIP ViT-L & & \cmark & \cmark && 54.4 & & 36.5 && 55.5 && - && - && 85.0 && 94.9\\
PAC-S~\cite{sarto2023positive} & CLIP ViT-L & & \cmark &  && 55.5 & & 36.8 && 52.4 && 47.9 && 56.5 && 82.2 && 91.9\\
RefPAC-S~\cite{sarto2023positive} & CLIP ViT-L & & \cmark & \cmark && {57.1} & & {37.7} && 55.5 && 50.4 && - && 85.0 && \textbf{95.3}\\
PAC-S++~\cite{sarto2024positive} & CLIP ViT-L & & \cmark & && 57.4 & & 38.5 && 53.6 && - && \textbf{62.0} && 82.4 && -\\
RefPAC-S++~\cite{sarto2024positive} & CLIP ViT-L & & \cmark & \cmark && \textbf{57.9} & & \textbf{\underline{38.8}} && \textbf{55.6} && - && 61.6 && 84.7 && -\\
\midrule
\rowcolor{TitleColor}\multicolumn{18}{l}{\textit{Learnable Supervised}} & \\
Polos~\cite{wada2024polos} & CLIP ViT-B & & \cmark & \cmark && 56.4 & & 37.8 && \textbf{\underline{57.8}} && 53.9 && 57.6 && {86.5} && 93.3\\
DENEB~\cite{matsuda2024deneb} & CLIP ViT-B & & \cmark & \cmark && 56.5 & & 38.0 && - && {54.1} && 57.9 && 87.0 && 95.1\\
DENEB~\cite{matsuda2024deneb} & CLIP ViT-L & & \cmark & \cmark && \textbf{56.8} & & \textbf{38.3} && - && \textbf{\underline{54.3}} && \textbf{58.2} && \textbf{\underline{87.8}} && \textbf{95.4}\\
\midrule
\rowcolor{TitleColor}\multicolumn{18}{l}{\textit{Learnable Fine-grained Oriented}} & \\
BRIDGE~\cite{sarto2024bridge} & CLIP ViT-B & & \cmark &  && 54.8 & & 36.1 && - && - && 55.0 && 82.6 && 91.5\\
BRIDGE~\cite{sarto2024bridge} & CLIP ViT-L & & \cmark &  && 55.8 & & 36.3 && - && - && 57.2 && 82.9 && 93.0\\
HICE-S~\cite{zeng2024hicescore} & SAM+Alpha-CLIP ViT-L & & \cmark &  && 56.4 & & 37.2 && - && - && 57.9 && 86.1 && 93.1\\
RefHICE-S~\cite{zeng2024hicescore} & SAM+Alpha-CLIP ViT-L & & \cmark & \cmark && {57.7} & & \textbf{38.2} && - && - && 58.7 && \textbf{87.3} && \textbf{96.4}\\
HiFi-S~\cite{yao2024hifi} & SAM+BLIP-2 & & \cmark &  && \textbf{\underline{58.4}} & & - && - && - && \textbf{\underline{65.8}} && 83.0 & & -\\
\midrule
\rowcolor{TitleColor}\multicolumn{18}{l}{\textit{LLMs-based}} & \\
CLAIR~\cite{chan2023clair} & GPT-3.5 & &  & \cmark && 48.3 & & - && - && \textbf{52.7} && 61.0 && 78.7 && 81.4\\
FLEUR~\cite{lee2024fleur} & LLaVA v1.5-13B & & \cmark &  && 53.0 & & {38.6} && - && - && {63.5} && 83.2 && {96.8} \\
RefFLEUR~\cite{lee2024fleur} & LLaVA v1.5-13B & & \cmark & \cmark && \textbf{51.9} & & \textbf{\underline{38.8}} && - && - &&  \textbf{64.2} && \textbf{85.5} && \textbf{\underline{97.3}}\\
\bottomrule
\end{tabular}
}
\vspace{-0.15cm}
\caption{A quantitative comparison between metrics, reporting correlation scores on Flickr8k-Expert, Flickr8k-CF, Polaris, Nebula, and Composite, along with accuracy on Pascal-50S and FOIL. The best scores within each category are in bold, overall best scores are underlined.}
\label{tab:others}
\vspace{-0.3cm}
\end{table*}

\tit{Fine-grained Oriented Metrics}
More recent metrics still utilize multimodal models
(\eg~CLIP) while integrating additional components to enhance performance, particularly by improving their ability to reward fine-grained details.

Some metrics adopt a reference-free approach, arguing that reliance on reference captions introduces challenges and additional costs, thereby increasing the complexity of the evaluation process. For instance, BRIDGE~\cite{sarto2024bridge} is a reference-free metric that employs a dual-encoder architecture combined with a mapping module responsible for generating multimodal pseudo-captions, which integrate textual and dense visual features. Similarly, InfoMetIC~\cite{hu2023infometic} is designed to provide fine-grained feedback on caption quality. Beyond assessing precision and recall, it identifies specific errors, such as incorrect words and unmentioned image regions. To achieve this, InfoMetIC incorporates a fusion module to model intra- and inter-modality relationships, along with a fine-grained module that enhances the accuracy of error localization in both text and image content.

Available in both reference-free and reference-based versions, HICE-S~\cite{zeng2024hicescore} highlights the limitations of CLIP-based metrics, which primarily assess global image-text compatibility but often struggle with detecting local textual hallucinations and maintaining sensitivity to small visual objects. To overcome these challenges, HICE-S introduces a hierarchical scoring mechanism that leverages the SAM model~\cite{kirillov2023segment} to generate masks for localized visual regions and corresponding textual phrases. These are then processed through a modified version of the CLIP architecture~\cite{sun2024alpha}, enhancing the model's ability to capture fine-grained visual-semantic relationships. 

Following a similar hierarchical approach, HiFi-S~\cite{yao2024hifi} is a fine-grained image description evaluation metric that represents both text and images as parsing graphs. These graphs organize multi-granular instances into a hierarchical structure based on their inclusion relationships, enabling a comprehensive scene analysis across modalities from global to local levels. Additionally, HiFi-S incorporates an LLM to evaluate the fluency of candidate descriptions, further enhancing the evaluation process.

\subsection{LLMs-based Metrics}
In recent years, the integration of LLMs into the captioning evaluation pipeline has gained popularity, as demonstrated by HiFi-S. While first attempts compared generated sentences with reference captions using BERT-based embeddings, more recent approaches leverage the advanced reasoning and extensive pre-training capabilities of LLMs to produce more robust evaluation scores.  
For example, CLAIR~\cite{chan2023clair} utilizes an LLM to rate the alignment of a candidate caption with a set of reference captions. Notably, CLAIR evaluates captions without considering image content. In response, FLEUR~\cite{lee2024fleur} is a reference-free, explainable evaluation metric for image captioning, which directly leverages a score generated by an MLLM and adjusted with a smoothing function to better align with human judgments.

\subsection{Hallucination-oriented Metrics}
In image captioning, hallucination refers to the inclusion of information, objects, or details in a generated caption that are not present in the corresponding image, compromising the reliability and quality of the description. Accurately identifying captions with potential object hallucinations is therefore essential. Metrics such as CHAIR~\cite{rohrbach2018object} and ALOHa~\cite{petryk2024aloha} are specifically designed to address this challenge.
Specifically, the CHAIR metric calculates what proportion of words generated is actually in the image according to the ground-truth sentences and detected object. However, this metric is restricted to a fixed set of COCO objects and their synonyms. To overcome this limitation, ALOHa introduces an open-vocabulary approach that utilizes LLMs to detect object hallucinations. Specifically, ALOHa extracts groundable objects from the candidate caption using an LLM, measures their semantic similarity to reference objects in captions or detections, and applies Hungarian matching to compute the final hallucination score.

%% file: sections/03_experiments.tex
\section{Experimental Evaluation}
This section presents a comprehensive analysis of captioning evaluation metrics through quantitative experiments. The evaluation spans multiple dimensions, including \hlpink{correlation with human judgments}, \hlblue{ranking accuracy}, and \hlpurple{sensitivity to object hallucinations}. Experiments are conducted on diverse datasets, with results summarized in Table~\ref{tab:others}.
Additionally, we assess how well existing metrics evaluate captions generated by modern MLLMs, with results presented in Table~\ref{tab:captioners}.

\subsection{Correlation with Human Judgment}
We analyze the correlation of metrics with human judgment, highlighting their varying behaviours. 
Experimental results are reported on standard captioning evaluation datasets, including Flickr8k-Expert, Flickr8k-CF, and Composite, along with the more recent Polaris and Nebula datasets.

\tit{Datasets}
Flickr8k-Expert~\cite{hodosh2013framing} contains 17k annotations for 5,664 images, with each pair scored from 1 (no correlation) to 4 (accurate depiction). Flickr8k-CF~\cite{hodosh2013framing} provides 145k binary quality judgments for 48k pairs across 1,000 unique images, using the mean proportion of ``yes'' votes as the alignment score. The Composite dataset~\cite{aditya2015images} includes 12k human ratings for image-caption pairs (with around 4k unique images), assessed on a 1–5 scale. However, these datasets lack model-generated captions, leading to a domain gap when applying them to train evaluation metrics.
To address this limitation, 
the Polaris dataset~\cite{wada2024polos} introduces 131k human judgments from 550 evaluators (\ie, around ten times larger than previous datasets) covering both human-written and machine-generated captions from ten image captioning models. Expanding on Polaris, the Nebula dataset~\cite{matsuda2024deneb} includes 32,978 images with human judgments from 805 annotators. In line with prior studies~\cite{hessel2021clipscore,wada2024polos}, we use Kendall’s correlation coefficient $\tau_b$ for Flickr8k-CF and Kendall’s $\tau_c$ for the other datasets.

\tit{Experimental Analysis and Key Takeaways}
Among rule-based methods, CIDEr demonstrates the highest performance across most datasets, with the exception of Flickr8k-Expert and Composite, where SPICE outperforms it by +1 and +2.6 points, respectively. 
Within the learnable metrics, RefPAC-S++ (ViT-L) achieves the best results on Flickr8k-Expert, Flickr8k-CF, and Polaris, while ViLBERT-S yields the highest scores among metrics evaluated on the Nebula dataset. Notably, reference-based metrics generally outperform their reference-free counterparts. However, among the reference-free methods, InfoMetIC and PAC-S variants demonstrate superior performance, emphasizing the importance of refining large pre-trained backbones in the absence of reference captions. When scaling the backbone size, PAC-S variants maintain superior performance, demonstrating their robustness and scalability. In a supervised setting, comparisons are more complex due to differences in backbone sizes. Overall, DENEB (ViT-L) achieves the highest results across multiple datasets. Interestingly, despite leveraging a supervised approach and incorporating an additional learnable module, DENEB is outperformed by RefPAC-S++ (ViT-L), which is trained in an unsupervised manner. 
This highlights the effectiveness of leveraging CLIP pre-training and enhancing it with regularization through additional generated visual-textual pairs.
For fine-grained evaluation, the HiFi metric, employing a hierarchical parsing graph, achieves the best results in a reference-free setting, surpassing HICE-S by +7.9 and BRIDGE (ViT-L) by +8.6 on the Composite dataset. On the Flickr8k-CF dataset, the best performance is instead achieved by RefHICE-S.
Among LLM-based metrics, FLEUR performs best, highlighting the importance of leveraging a multimodal approach that incorporates input images, unlike the CLAIR metric which relies solely on reference captions for evaluation. This underscores the critical role of visual information in the captioning task.

Overall, no single metric consistently outperforms all others across datasets. 
Interestingly, on the Flickr8k-CF dataset, comparable results are achieved using both RefPAC-S++ and the MLLM-based metric FLEUR. This comparison is particularly noteworthy, as the FLEUR metric, based on a LLaVA model with 13B parameters, achieves similar performance to metrics that utilize a smaller CLIP model with around 400M parameters. This highlights that for specific tasks like captioning evaluation, refining a pre-trained embedding space may be more effective than relying on a larger, multi-task embedding. Moreover, there is a significant performance gap between rule-based metrics and newer approaches, highlighting the importance of leveraging input images and the advantages of large-scale pre-training in achieving superior performance.

\subsection{Pairwise Ranking}
We focus on the pairwise ranking ability of current captioning metrics, measuring performance on the Pascal-50S dataset. 

\tit{Dataset} Pascal-50S~\cite{vedantam2015cider} evaluates captioning metrics using pairwise preference judgments. 
It consists of 4,000 sentence pairs linked to 1,000 images, each with 50 reference captions. Human judges determine which caption better describes the image, categorizing pairs into human-correct, human with one incorrect caption, human vs. machine-generated, and machine-generated. In this setting, we compute accuracy instead of correlation scores, reporting the averaged results across the four categories. 

\tit{Experimental Analysis and Key Takeaways}
Among rule-based metrics, METEOR achieves the highest accuracy, outperforming CIDEr by +1 point. For learnable unsupervised metrics, the results deviate from trends seen in correlation-based evaluations, where larger backbones and reference-based metrics typically excel. In this case, InfoMetIC, a reference-free metric using the ViT-B backbone, emerges as the best among all metrics. Its ability to identify incorrect semantic words and unmentioned visual regions proves advantageous for this task. Conversely, for other learnable and LLM-based metrics, the results closely align with those observed in human correlation evaluations, with the highest overall accuracy achieved by DENEB (ViT-L). 

\begin{table*}[t]
\small
\centering
\setlength{\tabcolsep}{.33em}
\resizebox{0.98\linewidth}{!}{
\begin{tabular}{ccc cc ccc ccc c ccc }
\toprule
& & & LLM & & Length & BLEU-4 & METEOR & CIDEr & CLIP-S & PAC-S++ & RefCLIP-S & RefPAC-S++ & Polos & BRIDGE \\
\midrule
\multirow{10}{*}{{\rotatebox[origin=c]{90}{\textit{Captioning Models}}}} 
& & Show and Tell~\cite{vinyals2015show} & - & & 9.1 & 31.4 & 25.0 & 97.2 & 0.715 & 0.654 & 0.779 & 0.752 & 0.585 & 0.788 \\
& & Show, Attend and Tell~\cite{xu2015show} & - & & 9.1 & 33.4 & 26.2 & 104.6 & 0.727 & 0.670 & 0.790 & 0.766 & 0.609 & 0.804 \\
& & Up-Down~\cite{anderson2018bottom} & - & & 9.5 & 36.7 & 27.9 & 122.7 & 0.740 & 0.680 & 0.804 & 0.778 & 0.640 & 0.821 \\
& & SGAE~\cite{yang2019auto} & - & & 9.4 & 38.6 & 28.8 & 129.8 & 0.750 & 0.691 & 0.812 & 0.786 & 0.655 & 0.833 \\
& & AoANet~\cite{huang2019attention} & - & & 9.5 & 39.1 & 29.0 & 128.9 & 0.753 & 0.693 & 0.813 & 0.787 & 0.660 & 0.836 \\
& & $\mathcal{M}^2$ Transformer~\cite{cornia2020meshed} & - & & 9.7 & 39.1 & 29.2 & 131.2 & 0.757 & 0.699 & 0.813 & 0.791 & 0.629 & 0.841 \\
& & X-Transformer~\cite{pan2020x} & - & & 9.6  & 39.7 & 29.5 & 132.8 & 0.762 & 0.701 & 0.819 & 0.792 & 0.668 & 0.845\\
& & VinVL~\cite{zhang2021vinvl} & - & & 10.0 & {41.0} & {31.1} & {140.9} & {0.784} & 0.715 & 0.836 & 0.805 & 0.708 & 0.865 \\
& & COS-Net~\cite{li2022comprehending} & - & & 9.6 & 42.0 & 30.6 & 141.1 & 0.773 & 0.712 & 0.829 & 0.803 & 0.692 & 0.859 \\
& & BLIP-2~\cite{li2023blip2} &  Flan T5-XL & & 9.6 & \textbf{43.8} & \textbf{31.7} & \textbf{146.0} & \textbf{0.782} & \textbf{0.719} & \textbf{0.838} & \textbf{0.810} & \textbf{0.716} & \textbf{0.868} \\
\midrule
\multirow{12}{*}{{\rotatebox[origin=c]{90}{\textit{General Purpose MLLMs}}}} 
 & & InstructBLIP~\cite{dai2023instructblip} & Vicuna-7B & & \cellcolor{myYellow}10.2 & \textbf{41.2} & \textbf{31.8} & \textbf{142.2} & 0.786 & {0.721} & 0.838 & \textbf{0.810} & \textbf{0.714} & 0.871 \\
& & LLaVA-1.5~\cite{liu2024improved} & Vicuna-7B  & & \cellcolor{myYellow}14.5 & 8.1 & 28.0 & 69.6 & 0.785 & 0.707 & 0.828 & 0.794 & 0.666 & 0.867 \\ 
& & IDEFICS~\cite{laurenccon2023obelisc} & Llama-7B & & \cellcolor{myYellow}8.8 & 36.8 & 28.3 & 125.1 & 0.788 & 0.711 & \textbf{0.840} & 0.803 & 0.699 & 0.863 \\
& & IDEFICS-2~\cite{laurenccon2024matters} & Mistral-7B & & \cellcolor{myYellow}12.2 & 19.1 & 24.3 & 74.1 & 0.800 & 0.711 & 0.819 & 0.780 & 0.626 & 0.865 \\
& & IDEFICS-3~\cite{laurenccon2024building} & Llama-3-8B & & \cellcolor{myYellow}14.2 & 17.4 & 24.5 & 62.2 & {0.801} & 0.710 & 0.811 & 0.771 & 0.615 & 0.865 \\
& & Llama-3.2~\cite{dubey2024llama} & Llama-3.2-11B & & \cellcolor{myYellow}22.2 & 14.3 & 25.8 & 46.0 & \textbf{0.827} & \textbf{0.734} & 0.828 & 0.796 & 0.692 & \textbf{0.900} \\
\cmidrule{3-15}
& & InstructBLIP~\cite{dai2023instructblip} & Vicuna-7B & & \cellcolor{myGreen}43.3 & 9.8 & \textbf{25.0} & 2.8 &{0.828} & \textbf{0.738} & 0.817 & \textbf{0.792} & \textbf{0.709} & \textbf{0.912} \\
& & LLaVA-1.5~\cite{liu2024improved} & Vicuna-7B & & \cellcolor{myGreen}49.5 & 8.3 & 23.0 & 0.3 & 0.813 & 0.722 & 0.808 & 0.783 & 0.689 & 0.898 \\
& & IDEFICS~\cite{laurenccon2023obelisc} & Llama-7B & & \cellcolor{myGreen}29.0 & \textbf{11.4} & 24.8 & \textbf{43.9} & 0.792 & {0.719} & 0.815 & 0.788 & 0.679 & 0.867 \\
& & IDEFICS-2~\cite{laurenccon2024matters} & Mistral-7B & & \cellcolor{myGreen}22.1  & 7.3 & 20.0 & 31.7 & 0.728 & 0.683 & 0.773 & 0.760 & 0.575 & 0.794 \\
& & IDEFICS-3~\cite{laurenccon2024building} & Llama-3-8B & & \cellcolor{myGreen}123.6 & 2.0 & 12.2 & 8.3 & 0.777 & 0.697 & 0.770 & 0.748 & 0.605 & 0.856 \\ 
& & Llama-3.2~\cite{dubey2024llama} & Llama-3.2-11B & & \cellcolor{myGreen}31.4 & 10.2 & 23.9 & 30.5 & \textbf{0.832} & 0.736 & \textbf{0.818} & 0.790 & 0.689 & 0.906 \\
\midrule 
\rowcolor{TitleColor}
& & \textit{Humans} & & & \textit{10.4} & - & \textit{24.1} & \textit{87.6} & {\textit{0.782}} & \textit{0.710} & \textit{0.822} & \textit{0.792} & \textit{0.654} & \textit{0.856} \\
\bottomrule
\end{tabular}
}
\vspace{-0.15cm}
\caption{Evaluation scores on the COCO test set comparing traditional captioning models with general-purpose MLLMs. For MLLMs, we assess both short captions and longer, unconstrained descriptions generated using the default prompt of the model.}
\label{tab:captioners}
\vspace{-0.3cm}
\end{table*}

\subsection{Sensitivity to Object Hallucinations}
We evaluate robustness to hallucinations on the FOIL dataset.

\tit{Dataset} FOIL~\cite{shekhar2017foil} consists of image-caption pairs derived from COCO~\cite{lin2014microsoft}, where captions are intentionally altered by introducing a single erroneous word, that makes the modified caption closely resembles the original but includes a specific mistake. In our evaluation, we compute the percentage of times the original caption gets the highest score, according to each metric. 

\tit{Experimental Analysis and Key Takeaways} In this task, CIDEr outperforms all other rule-based metrics by a substantial margin, with a gain of around +10 points. Among learnable metrics, RefHICE achieves the highest accuracy, followed by DENEB and RefPAC-S, both using the ViT-L backbone. Reference-free versions of HICE and PAC-S experience a significant performance drop, indicating the need for reference captions to detect hallucinated objects.
The results also highlight the advantages of stronger backbones. For instance, PAC-S (ViT-L) outperforms PAC-S (ViT-B) by +2 points, demonstrating the impact of model architecture on performance. Notably, rule-based metrics like CIDEr are significantly outperformed by modern multimodal approaches, with a performance gap of approximately 10 points compared to RefPAC-S (ViT-L) and nearly 12 points against the LLM-based RefFLEUR. This substantial disparity highlights the limitations of rule-based approaches in effectively capturing the nuanced semantics of image captions, particularly in detecting subtle errors such as hallucinations. Metrics like RefPAC-S and RefFLEUR benefit not only from powerful backbones but also from their ability to align text with visual content, allowing them to detect discrepancies much more accurately than traditional metrics. These results empathize the growing need 
for evaluation metrics that assess not just linguistic fluency but also semantic accuracy and visual relevance through advanced multimodal understanding.

\subsection{System-level Correlation}
The key objective of this study is to assess whether well-established captioning metrics effectively evaluate captions generated by modern MLLMs, especially when their style deviates from traditional COCO captions. 
The potential limitations of these metrics originate from their design: rule-based metrics rely on reference captions that closely adhere to COCO-style annotations, while learnable metrics are primarily fine-tuned on COCO or similar datasets. As a result, their ability to accurately evaluate captions with diverse linguistic structures or lengths remains uncertain.
To investigate this, we assess MLLM-generated captions in two formats: \hlyellow{short captions}, comparable in length to COCO ones, and \hlgreen{longer, unconstrained descriptions}\footnote{For short captions, we use the prompt ``\texttt{Briefly describe the image}''. For longer ones, we rely on the MLLM default prompt (\eg~``\texttt{What is the content of the image?}'').}.
This enables us to analyze metric performance across varying caption styles and lengths.

In Table~\ref{tab:captioners}, we evaluate popular image description models on the COCO test set using a range of metrics to assess their effectiveness and highlight differences in their behavior\footnote{Note that for BLIP-2 we employ the captioning-specific model fine-tuned on COCO image-caption pairs.}. Our analysis includes both traditional captioning models as well as modern MLLMs designed for broader tasks. Additionally, we establish a human-based baseline by randomly selecting one human-annotated caption from the five provided in COCO and comparing it against the remaining references\footnote{BLEU is not reported for the human-based baseline as its value is sensitive to the number of reference captions.}. 
For evaluation, we employ a diverse set of metrics, including standard rule-based scores such as BLEU, METEOR and CIDEr, as well as two widely used learnable metrics, CLIP-S and PAC-S++, along with their reference-based counterparts. Additionally, we include two recent evaluation methods: Polos, a supervised metric, and BRIDGE, which focuses on fine-grained visual details. This selection allows for a comprehensive assessment of how different evaluation methods capture caption quality across various models.

\begin{figure*}[t]
\centering
\includegraphics[width=0.99\linewidth]{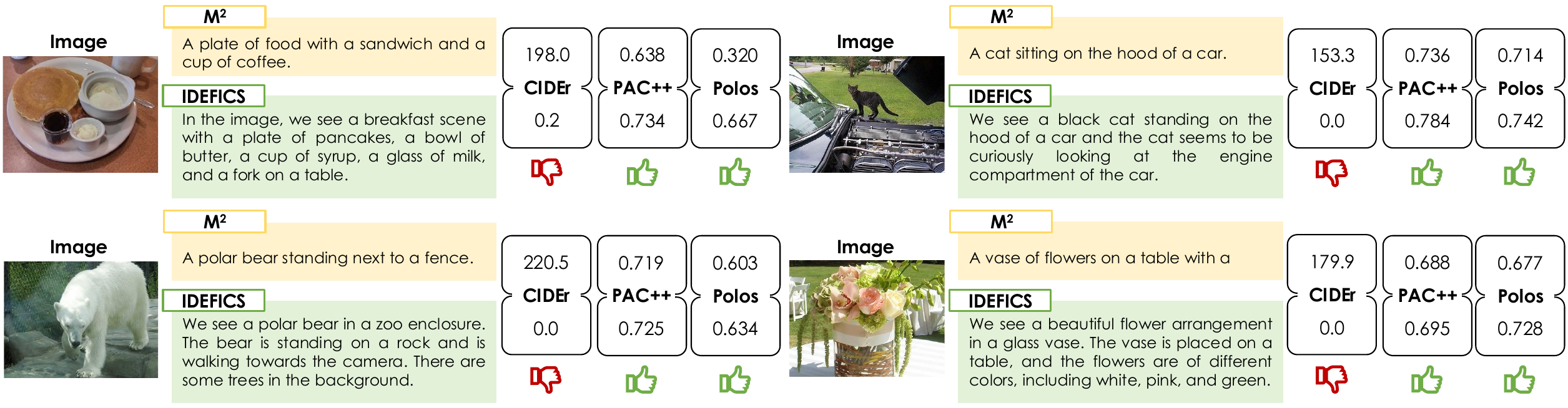}
\vspace{-0.15cm}
\caption{Qualitative examples showing the differences between captions generated by a traditional captioning model (\ie, $\mathcal{M}^2$ Transformer) and MLLM-style captions (\ie, IDEFICS). Length and style variations affect rule-based metrics, while learnable metrics remain robust.
}
\label{fig:qualitative}
\vspace{-.3cm}
\end{figure*}

As it can be seen, for models explicitly trained for the image captioning task, such as $\mathcal{M}^2$ Transformer, the generated captions tend to align with COCO in both length and style. Under these conditions, traditional evaluation metrics like CIDEr, effectively recognize the superior quality of BLIP-2 captions. Similarly, recently developed metrics leveraging large-scale pre-trained models maintain strong performance.

When evaluating captions generated by MLLMs interesting patterns emerge. If the captions remain similar in length to the ones contained in COCO, rule-based metrics still perform effectively: CIDEr, for instance, assigns a score of 142.2 to captions generated by InstructBLIP. However, as caption length increases, these metrics exhibit a sharp decline in scores. Notably, when evaluating LLaVA-1.5, which produces significantly longer captions than COCO ones, CIDEr score drops dramatically to just 0.3. Qualitative results showing this trend are reported in Fig.~\ref{fig:qualitative}.
This decline is primarily due to the reliance of rule-based metrics on COCO-style reference captions. When comparing longer and more detailed MLLM-generated captions against these shorter and simpler references, these metrics often struggle to compute realistic scores, leading to misleading evaluations.

In contrast, more recent metrics that incorporate both large-scale pre-training and visual input exhibit greater robustness to variations in caption length. A slight decline in performance is still observed: for example, PAC-S++ decreases from 0.710 to 0.697 when caption length increases nearly ninefold (\ie, for IDEFICS-3). This reduction primarily reflects a lower confidence level rather than an outright failure of the metric. 
In fact, although these metrics were not explicitly designed for evaluating long, highly detailed captions, they still maintain a high degree of reliability.

For learnable reference-based metrics such as RefCLIP-S, RefPAC-S++, and Polos, performance degrades more significantly compared to their reference-free counterparts, reflecting a trend similar to that observed in rule-based metrics. However, unlike rule-based solutions, these metrics retain their ability to distinguish high-quality captions, correctly rewarding models such as Llama-3.2 and InstructBLIP.

Overall, these results indicate that recent learnable metrics designed for image captioning remain valuable for evaluating longer and more detailed captions generated by MLLMs, demonstrating reliability despite variations in length and style.
However, reference-free metrics are generally preferable, as they more effectively distinguish high-quality captions and exhibit greater robustness to variations in length.

%% file: sections/04_future_directions.tex
\section{Future Directions and Open Problems}\label{sec:future}

Despite progress in image captioning evaluation, several challenges remain, offering opportunities for future research.

\tit{Benchmark Evolution} 
Traditional evaluation datasets mainly feature short captions similar to those in COCO, creating a gap with the longer, more detailed outputs of modern MLLMs. While recent efforts, like Polaris and Nebula, incorporate captions from standard models, they often maintain the concise style of COCO descriptions.
A valuable direction would involve creating new benchmarks specifically designed to assess the quality of metrics on longer, richer captions that better reflect the MLLM outputs, addressing challenges like synonyms, paraphrases, and domain-specific terms.m
\tit{Explainability in Metrics} Most evaluation metrics generate scores without offering insights into the rationale behind their assessments, limiting their usefulness for improving captioning models. A few metrics, such as CLAIR, have advanced in this area by leveraging the interpretative capabilities of LLMs to offer explanations alongside scores. Building upon these advancements, future research should focus on further enhancing explainability, facilitating a deeper understanding of the strengths and weaknesses of evaluation models.

\tit{Detecting Hallucinations}
Hallucination remains a major challenge in captioning literature, with models often generating inaccurate or non-existent details. While benchmarks like FOIL help identify simpler errors, modern MLLMs produce more complex hallucinations, such as distorted relationships or entirely invented elements. To address this, more diverse datasets are needed to effectively assess these complex errors, including test cases specifically designed to challenge evaluation metrics on altered attributes, incorrect relationships, or non-existent entities and actions. Additionally, although some metrics targeting hallucinations have emerged, more advanced solutions are required to capture the diversity and semantic richness of MLLM-generated captions. 

\tit{Metrics Personalization}
Current evaluation metrics rely on reference captions or image-caption alignment, but they fail to accommodate the diversity of user preferences or task-specific requirements.
Future research should focus on the development of personalized evaluation metrics, that can let users prioritize factors such as detail, brevity, stylistic preferences, or domain relevance.
By incorporating these custom priorities, evaluation can become more adaptable and meaningful across a variety of applications and user needs.

%% file: sections/05_conclusions.tex
\section{Conclusion}\label{sec:conclusions}
This survey examines image captioning evaluation metrics with the advent of MLLMs, comparing their performance on standard datasets and analyzing their adaptability to evolving caption styles. Our analysis reveals that traditional rule-based metrics struggle to evaluate longer, more detailed captions, while learnable metrics remain effective. Moreover, we also highlight key open challenges and promising future directions, emphasizing the need for more robust, context-aware evaluation frameworks that can effectively assess the growing diversity and complexity of image captioning models.

%% file: ijcai25.bbl
\begin{thebibliography}{}

\bibitem[\protect\citeauthoryear{Aditya \bgroup \em et al.\egroup }{2015}]{aditya2015images}
Somak Aditya, Yezhou Yang, Chitta Baral, Cornelia Fermuller, et~al.
\newblock {From Images to Sentences through Scene Description Graphs using Commonsense Reasoning and Knowledge}.
\newblock {\em arXiv:1511.03292}, 2015.

\bibitem[\protect\citeauthoryear{Agrawal \bgroup \em et al.\egroup }{2019}]{agrawal2019nocaps}
Harsh Agrawal, Karan Desai, Xinlei Chen, Rishabh Jain, Dhruv Batra, Devi Parikh, et~al.
\newblock nocaps: novel object captioning at scale.
\newblock In {\em ICCV}, 2019.

\bibitem[\protect\citeauthoryear{Anderson \bgroup \em et al.\egroup }{2016}]{spice2016}
Peter Anderson, Basura Fernando, Mark Johnson, and Stephen Gould.
\newblock {SPICE: Semantic Propositional Image Caption Evaluation}.
\newblock In {\em ECCV}, 2016.

\bibitem[\protect\citeauthoryear{Anderson \bgroup \em et al.\egroup }{2018}]{anderson2018bottom}
Peter Anderson, Xiaodong He, Chris Buehler, Damien Teney, et~al.
\newblock {Bottom-Up and Top-Down Attention for Image Captioning and Visual Question Answering}.
\newblock In {\em CVPR}, 2018.

\bibitem[\protect\citeauthoryear{Banerjee and Lavie}{2005}]{banerjee2005meteor}
Satanjeev Banerjee and Alon Lavie.
\newblock {METEOR: An automatic metric for MT evaluation with improved correlation with human judgments}.
\newblock In {\em ACL Workshops}, 2005.

\bibitem[\protect\citeauthoryear{Chan \bgroup \em et al.\egroup }{2023}]{chan2023clair}
David Chan, Suzanne Petryk, Joseph~E Gonzalez, Trevor Darrell, et~al.
\newblock {CLAIR: Evaluating Image Captions with Large Language Models}.
\newblock In {\em EMNLP}, 2023.

\bibitem[\protect\citeauthoryear{Chen \bgroup \em et al.\egroup }{2020}]{chen2020uniter}
Yen-Chun Chen, Linjie Li, Licheng Yu, Ahmed El~Kholy, Faisal Ahmed, Zhe Gan, Yu~Cheng, and Jingjing Liu.
\newblock {UNITER: UNiversal Image-TExt Representation Learning}.
\newblock In {\em ECCV}, 2020.

\bibitem[\protect\citeauthoryear{Cornia \bgroup \em et al.\egroup }{2020}]{cornia2020meshed}
Marcella Cornia, Matteo Stefanini, Lorenzo Baraldi, and Rita Cucchiara.
\newblock {Meshed-Memory Transformer for Image Captioning}.
\newblock In {\em CVPR}, 2020.

\bibitem[\protect\citeauthoryear{Dai \bgroup \em et al.\egroup }{2023}]{dai2023instructblip}
Wenliang Dai, Junnan Li, Dongxu Li, Anthony Meng~Huat Tiong, Junqi Zhao, et~al.
\newblock {InstructBLIP: Towards General-purpose Vision-Language Models with Instruction Tuning}.
\newblock {\em arXiv:2305.06500}, 2023.

\bibitem[\protect\citeauthoryear{Devlin \bgroup \em et al.\egroup }{2018}]{devlin2018bert}
Jacob Devlin, Ming-Wei Chang, Kenton Lee, and Kristina Toutanova.
\newblock {BERT: Pre-training of Deep Bidirectional Transformers for Language Understanding}.
\newblock In {\em NAACL}, 2018.

\bibitem[\protect\citeauthoryear{Dosovitskiy \bgroup \em et al.\egroup }{2021}]{dosovitskiy2020image}
Alexey Dosovitskiy, Lucas Beyer, Alexander Kolesnikov, Dirk Weissenborn, Xiaohua Zhai, et~al.
\newblock {An Image is Worth 16x16 Words: Transformers for Image Recognition at Scale}.
\newblock In {\em ICLR}, 2021.

\bibitem[\protect\citeauthoryear{Dubey \bgroup \em et al.\egroup }{2024}]{dubey2024llama}
Abhimanyu Dubey, Abhinav Jauhri, Abhinav Pandey, Abhishek Kadian, Ahmad Al-Dahle, Aiesha Letman, Akhil Mathur, et~al.
\newblock {The Llama 3 Herd of Models}.
\newblock {\em arXiv:2407.21783}, 2024.

\bibitem[\protect\citeauthoryear{Hessel \bgroup \em et al.\egroup }{2021}]{hessel2021clipscore}
Jack Hessel, Ari Holtzman, Maxwell Forbes, Ronan~Le Bras, and Yejin Choi.
\newblock {CLIPScore: A Reference-free Evaluation Metric for Image Captioning}.
\newblock In {\em EMNLP}, 2021.

\bibitem[\protect\citeauthoryear{Hodosh \bgroup \em et al.\egroup }{2013}]{hodosh2013framing}
Micah Hodosh, Peter Young, and Julia Hockenmaier.
\newblock {Framing Image Description as a Ranking Task: Data, Models and Evaluation Metrics}.
\newblock {\em JAIR}, 47:853--899, 2013.

\bibitem[\protect\citeauthoryear{Hu \bgroup \em et al.\egroup }{2023}]{hu2023infometic}
Anwen Hu, Shizhe Chen, Liang Zhang, and Qin Jin.
\newblock {InfoMetIC: An Informative Metric for Reference-free Image Caption Evaluation}.
\newblock In {\em ACL}, 2023.

\bibitem[\protect\citeauthoryear{Huang \bgroup \em et al.\egroup }{2019}]{huang2019attention}
Lun Huang, Wenmin Wang, Jie Chen, and Xiao-Yong Wei.
\newblock {Attention on Attention for Image Captioning}.
\newblock In {\em ICCV}, 2019.

\bibitem[\protect\citeauthoryear{Jiang \bgroup \em et al.\egroup }{2019}]{jiang2019tiger}
Ming Jiang, Qiuyuan Huang, Lei Zhang, Xin Wang, et~al.
\newblock {TIGEr: Text-to-Image Grounding for Image Caption Evaluation}.
\newblock In {\em EMNLP}, 2019.

\bibitem[\protect\citeauthoryear{Kim \bgroup \em et al.\egroup }{2022}]{kim2022mutual}
Jin-Hwa Kim, Yunji Kim, Jiyoung Lee, Kang~Min Yoo, and Sang-Woo Lee.
\newblock {Mutual Information Divergence: A Unified Metric for Multimodal Generative Models}.
\newblock In {\em NeurIPS}, 2022.

\bibitem[\protect\citeauthoryear{Kirillov \bgroup \em et al.\egroup }{2023}]{kirillov2023segment}
Alexander Kirillov, Eric Mintun, Nikhila Ravi, Hanzi Mao, Chloe Rolland, Laura Gustafson, Tete Xiao, Spencer Whitehead, Alexander~C Berg, et~al.
\newblock {Segment Anything}.
\newblock In {\em ICCV}, 2023.

\bibitem[\protect\citeauthoryear{Lauren{\c{c}}on \bgroup \em et al.\egroup }{2023}]{laurenccon2023obelisc}
Hugo Lauren{\c{c}}on, Lucile Saulnier, L{\'e}o Tronchon, Stas Bekman, Amanpreet Singh, et~al.
\newblock {OBELICS: An Open Web-Scale Filtered Dataset of Interleaved Image-Text Documents}.
\newblock In {\em NeurIPS}, 2023.

\bibitem[\protect\citeauthoryear{Lauren{\c{c}}on \bgroup \em et al.\egroup }{2024a}]{laurenccon2024building}
Hugo Lauren{\c{c}}on, Andr{\'e}s Marafioti, Victor Sanh, and L{\'e}o Tronchon.
\newblock {Building and better understanding vision-language models: insights and future directions}.
\newblock In {\em NeurIPS Workshops}, 2024.

\bibitem[\protect\citeauthoryear{Lauren{\c{c}}on \bgroup \em et al.\egroup }{2024b}]{laurenccon2024matters}
Hugo Lauren{\c{c}}on, L{\'e}o Tronchon, Matthieu Cord, and Victor Sanh.
\newblock {What matters when building vision-language models?}
\newblock In {\em NeurIPS}, 2024.

\bibitem[\protect\citeauthoryear{Lee \bgroup \em et al.\egroup }{2018}]{lee2018stacked}
Kuang-Huei Lee, Xi~Chen, Gang Hua, Houdong Hu, and Xiaodong He.
\newblock {Stacked Cross Attention for Image-Text Matching}.
\newblock In {\em ECCV}, 2018.

\bibitem[\protect\citeauthoryear{Lee \bgroup \em et al.\egroup }{2020}]{lee2020vilbertscore}
Hwanhee Lee, Seunghyun Yoon, Franck Dernoncourt, Doo~Soon Kim, Trung Bui, et~al.
\newblock {ViLBERTScore: Evaluating Image Caption Using Vision-and-Language BERT}.
\newblock In {\em EMNLP Workshops}, 2020.

\bibitem[\protect\citeauthoryear{Lee \bgroup \em et al.\egroup }{2021}]{lee2021umic}
Hwanhee Lee, Seunghyun Yoon, Franck Dernoncourt, Trung Bui, and Kyomin Jung.
\newblock {UMIC: An Unreferenced Metric for Image Captioning via Contrastive Learning}.
\newblock In {\em ACL}, 2021.

\bibitem[\protect\citeauthoryear{Lee \bgroup \em et al.\egroup }{2024}]{lee2024fleur}
Yebin Lee, Imseong Park, and Myungjoo Kang.
\newblock {FLEUR: An Explainable Reference-Free Evaluation Metric for Image Captioning Using a Large Multimodal Model}.
\newblock In {\em ACL}, 2024.

\bibitem[\protect\citeauthoryear{Li \bgroup \em et al.\egroup }{2022}]{li2022comprehending}
Yehao Li, Yingwei Pan, Ting Yao, and Tao Mei.
\newblock {Comprehending and Ordering Semantics for Image Captioning}.
\newblock In {\em CVPR}, 2022.

\bibitem[\protect\citeauthoryear{Li \bgroup \em et al.\egroup }{2023}]{li2023blip2}
Junnan Li, Dongxu Li, Silvio Savarese, and Steven Hoi.
\newblock {BLIP-2: Bootstrapping Language-Image Pre-training with Frozen Image Encoders and Large Language Models}.
\newblock In {\em ICML}, 2023.

\bibitem[\protect\citeauthoryear{Lin \bgroup \em et al.\egroup }{2014}]{lin2014microsoft}
Tsung-Yi Lin, Michael Maire, Serge Belongie, James Hays, Pietro Perona, et~al.
\newblock {Microsoft COCO: Common Objects in Context}.
\newblock In {\em ECCV}, 2014.

\bibitem[\protect\citeauthoryear{Lin}{2004}]{lin2004rouge}
Chin-Yew Lin.
\newblock Rouge: A package for automatic evaluation of summaries.
\newblock In {\em ACL Workshops}, 2004.

\bibitem[\protect\citeauthoryear{Liu \bgroup \em et al.\egroup }{2024}]{liu2024improved}
Haotian Liu, Chunyuan Li, Yuheng Li, and Yong~Jae Lee.
\newblock {Improved Baselines with Visual Instruction Tuning}.
\newblock In {\em CVPR}, 2024.

\bibitem[\protect\citeauthoryear{Lu \bgroup \em et al.\egroup }{2019}]{lu2019vilbert}
Jiasen Lu, Dhruv Batra, Devi Parikh, and Stefan Lee.
\newblock {ViLBERT: Pretraining Task-Agnostic Visiolinguistic Representations for Vision-and-Language Tasks}.
\newblock In {\em NeurIPS}, 2019.

\bibitem[\protect\citeauthoryear{Matsuda \bgroup \em et al.\egroup }{2024}]{matsuda2024deneb}
Kazuki Matsuda, Yuiga Wada, and Komei Sugiura.
\newblock {DENEB: A Hallucination-Robust Automatic Evaluation Metric for Image Captioning}.
\newblock In {\em ACCV}, 2024.

\bibitem[\protect\citeauthoryear{Pan \bgroup \em et al.\egroup }{2020}]{pan2020x}
Yingwei Pan, Ting Yao, Yehao Li, and Tao Mei.
\newblock {X-Linear Attention Networks for Image Captioning}.
\newblock In {\em CVPR}, 2020.

\bibitem[\protect\citeauthoryear{Papineni \bgroup \em et al.\egroup }{2002}]{papineni2002bleu}
Kishore Papineni, Salim Roukos, Todd Ward, and Wei-Jing Zhu.
\newblock {BLEU: a method for automatic evaluation of machine translation}.
\newblock In {\em ACL}, 2002.

\bibitem[\protect\citeauthoryear{Petryk \bgroup \em et al.\egroup }{2024}]{petryk2024aloha}
Suzanne Petryk, David~M Chan, Anish Kachinthaya, Haodi Zou, et~al.
\newblock {ALOHa: A New Measure for Hallucination in Captioning Models}.
\newblock In {\em NAACL}, 2024.

\bibitem[\protect\citeauthoryear{Radford \bgroup \em et al.\egroup }{2021}]{radford2021learning}
Alec Radford, Jong~Wook Kim, Chris Hallacy, Aditya Ramesh, Gabriel Goh, Sandhini Agarwal, Girish Sastry, et~al.
\newblock {Learning Transferable Visual Models From Natural Language Supervision}.
\newblock In {\em ICML}, 2021.

\bibitem[\protect\citeauthoryear{Rennie \bgroup \em et al.\egroup }{2017}]{rennie2017self}
Steven~J Rennie, Etienne Marcheret, Youssef Mroueh, Jarret Ross, et~al.
\newblock {Self-Critical Sequence Training for Image Captioning}.
\newblock In {\em CVPR}, 2017.

\bibitem[\protect\citeauthoryear{Rohrbach \bgroup \em et al.\egroup }{2018}]{rohrbach2018object}
Anna Rohrbach, Lisa~Anne Hendricks, Kaylee Burns, Trevor Darrell, et~al.
\newblock {Object Hallucination in Image Captioning}.
\newblock In {\em EMNLP}, 2018.

\bibitem[\protect\citeauthoryear{Sarto \bgroup \em et al.\egroup }{2023}]{sarto2023positive}
Sara Sarto, Manuele Barraco, Marcella Cornia, Lorenzo Baraldi, and Rita Cucchiara.
\newblock {Positive-Augmented Contrastive Learning for Image and Video Captioning Evaluation}.
\newblock In {\em CVPR}, 2023.

\bibitem[\protect\citeauthoryear{Sarto \bgroup \em et al.\egroup }{2024a}]{sarto2024bridge}
Sara Sarto, Marcella Cornia, Lorenzo Baraldi, and Rita Cucchiara.
\newblock {BRIDGE: Bridging Gaps in Image Captioning Evaluation with Stronger Visual Cues}.
\newblock In {\em ECCV}, 2024.

\bibitem[\protect\citeauthoryear{Sarto \bgroup \em et al.\egroup }{2024b}]{sarto2024positive}
Sara Sarto, Nicholas Moratelli, Marcella Cornia, Lorenzo Baraldi, et~al.
\newblock {Positive-Augmented Contrastive Learning for Vision-and-Language Evaluation and Training}.
\newblock {\em arXiv:2410.07336}, 2024.

\bibitem[\protect\citeauthoryear{Sharma \bgroup \em et al.\egroup }{2018}]{sharma2018conceptual}
Piyush Sharma, Nan Ding, Sebastian Goodman, and Radu Soricut.
\newblock {Conceptual Captions: A Cleaned, Hypernymed, Image Alt-text Dataset For Automatic Image Captioning}.
\newblock In {\em ACL}, 2018.

\bibitem[\protect\citeauthoryear{Shekhar \bgroup \em et al.\egroup }{2017}]{shekhar2017foil}
Ravi Shekhar, Sandro Pezzelle, Yauhen Klimovich, Aur{\'e}lie Herbelot, Moin Nabi, Enver Sangineto, et~al.
\newblock {FOIL it! Find One mismatch between Image and Language caption}.
\newblock In {\em ACL}, 2017.

\bibitem[\protect\citeauthoryear{Stefanini \bgroup \em et al.\egroup }{2022}]{stefanini2022show}
Matteo Stefanini, Marcella Cornia, Lorenzo Baraldi, Silvia Cascianelli, Giuseppe Fiameni, and Rita Cucchiara.
\newblock {From Show to Tell: A Survey on Deep Learning-based Image Captioning}.
\newblock {\em IEEE Trans. PAMI}, 45(1):539--559, 2022.

\bibitem[\protect\citeauthoryear{Sun \bgroup \em et al.\egroup }{2024}]{sun2024alpha}
Zeyi Sun, Ye~Fang, Tong Wu, Pan Zhang, Yuhang Zang, Shu Kong, et~al.
\newblock {Alpha-CLIP: A CLIP Model Focusing on Wherever You Want}.
\newblock In {\em CVPR}, 2024.

\bibitem[\protect\citeauthoryear{Vaswani \bgroup \em et al.\egroup }{2017}]{vaswani2017attention}
Ashish Vaswani, Noam Shazeer, Niki Parmar, Jakob Uszkoreit, Llion Jones, Aidan~N Gomez, et~al.
\newblock Attention is all you need.
\newblock In {\em NeurIPS}, 2017.

\bibitem[\protect\citeauthoryear{Vedantam \bgroup \em et al.\egroup }{2015}]{vedantam2015cider}
Ramakrishna Vedantam, Lawrence Zitnick, and Devi Parikh.
\newblock {CIDEr: Consensus-based Image Description Evaluation}.
\newblock In {\em CVPR}, 2015.

\bibitem[\protect\citeauthoryear{Vinyals \bgroup \em et al.\egroup }{2015}]{vinyals2015show}
Oriol Vinyals, Alexander Toshev, Samy Bengio, and Dumitru Erhan.
\newblock Show and tell: A neural image caption generator.
\newblock In {\em CVPR}, 2015.

\bibitem[\protect\citeauthoryear{Wada \bgroup \em et al.\egroup }{2024}]{wada2024polos}
Yuiga Wada, Kanta Kaneda, Daichi Saito, and Komei Sugiura.
\newblock {Polos: Multimodal Metric Learning from Human Feedback for Image Captioning}.
\newblock In {\em CVPR}, 2024.

\bibitem[\protect\citeauthoryear{Wang \bgroup \em et al.\egroup }{2021}]{wang2021faier}
Sijin Wang, Ziwei Yao, Ruiping Wang, Zhongqin Wu, and Xilin Chen.
\newblock {FAIEr: Fidelity and Adequacy Ensured Image Caption Evaluation}.
\newblock In {\em CVPR}, 2021.

\bibitem[\protect\citeauthoryear{Xu \bgroup \em et al.\egroup }{2015}]{xu2015show}
Kelvin Xu, Jimmy Ba, Ryan Kiros, Kyunghyun Cho, Aaron Courville, Ruslan Salakhutdinov, et~al.
\newblock {Show, Attend and Tell: Neural Image Caption Generation with Visual Attention}.
\newblock In {\em ICML}, 2015.

\bibitem[\protect\citeauthoryear{Yang \bgroup \em et al.\egroup }{2019}]{yang2019auto}
Xu~Yang, Kaihua Tang, Hanwang Zhang, and Jianfei Cai.
\newblock {Auto-Encoding Scene Graphs for Image Captioning}.
\newblock In {\em CVPR}, 2019.

\bibitem[\protect\citeauthoryear{Yao \bgroup \em et al.\egroup }{2024}]{yao2024hifi}
Ziwei Yao, Ruiping Wang, and Xilin Chen.
\newblock {HiFi-Score: Fine-Grained Image Description Evaluation with Hierarchical Parsing Graphs}.
\newblock In {\em ECCV}, 2024.

\bibitem[\protect\citeauthoryear{Yi \bgroup \em et al.\egroup }{2020}]{yi2020improving}
Yanzhi Yi, Hangyu Deng, and Jinglu Hu.
\newblock {Improving Image Captioning Evaluation by Considering Inter References Variance}.
\newblock In {\em ACL}, 2020.

\bibitem[\protect\citeauthoryear{Zeng \bgroup \em et al.\egroup }{2024}]{zeng2024hicescore}
Zequn Zeng, Jianqiao Sun, Hao Zhang, Tiansheng Wen, Yudi Su, Yan Xie, Zhengjue Wang, and Bo~Chen.
\newblock {HICEScore: A Hierarchical Metric for Image Captioning Evaluation}.
\newblock In {\em ACM Multimedia}, 2024.

\bibitem[\protect\citeauthoryear{Zhang \bgroup \em et al.\egroup }{2020}]{zhang2019bertscore}
Tianyi Zhang, Varsha Kishore, Felix Wu, Kilian~Q Weinberger, and Yoav Artzi.
\newblock {BERTScore: Evaluating Text Generation with BERT}.
\newblock In {\em ICLR}, 2020.

\bibitem[\protect\citeauthoryear{Zhang \bgroup \em et al.\egroup }{2021}]{zhang2021vinvl}
Pengchuan Zhang, Xiujun Li, Xiaowei Hu, Jianwei Yang, et~al.
\newblock {VinVL: Revisiting Visual Representations in Vision-Language Models}.
\newblock In {\em CVPR}, 2021.

\end{thebibliography}
